\documentclass[sigconf,onymous]{acmart}
\AtBeginDocument{%
  }

\setcopyright{acmlicensed}
\copyrightyear{2018}
\acmYear{2018}
\acmDOI{XXXXXXX.XXXXXXX}
\acmConference[Conference acronym 'XX]{Make sure to enter the correct
  conference title from your rights confirmation email}{June 03--05,
  2018}{Woodstock, NY}
\acmISBN{978-1-4503-XXXX-X/2018/06}
\usepackage{booktabs}
\usepackage{multirow}
\usepackage{array}
\usepackage{colortbl}
\usepackage{graphicx}
\usepackage{amsmath,amsfonts,amsthm}
\usepackage{pifont}
\usepackage{epstopdf}
\usepackage{xcolor}
\usepackage{subfigure}
\usepackage{dblfloatfix}
\usepackage{enumerate}
\usepackage{dsfont}
\usepackage{bm}
\usepackage{bbm}
\usepackage{makecell}
\usepackage{rotating}
\usepackage{url}
\usepackage{textcomp}
\usepackage{xspace}
\usepackage{float}
\usepackage{hyperref}
\definecolor{mygray}{RGB}{232,232,232}
\definecolor{myblue}{HTML}{1E90FF}
\definecolor{sectiongray}{RGB}{232,232,232}
\definecolor{blockgray}{RGB}{247,247,247}

\definecolor{champion}{RGB}{200,230,200}   
\definecolor{runnerup}{RGB}{255,255,200}   

\newcommand{\sys}{\textsc{ChronoQG}\xspace}
\acmSubmissionID{892}

\begin{document}

\title{ChronoQG: Towards a Temporally Expressive and Hop-Bounded Benchmark for Temporal Knowledge Graph Question Generation}

\author{Xuemeng Liu}
\authornote{Both authors contributed equally to this research.}
\affiliation{%
  \institution{Nankai University}
  \city{Tianjin}
  \country{China}
}
\email{xuemeng.liu@mail.nankai.edu.cn}

\author{Zhengpin Li}
\authornotemark[1]
\affiliation{%
  \institution{Peking University}
  \city{Beijing}
  \country{China}}
\email{zpli@pku.edu.cn}

\author{Wanpeng Tang}
\affiliation{%
  \institution{University of Electronic Science and Technology of China}
  \city{Chengdu}
  \country{China}}
\email{2023090910014@std.uestc.edu.cn}

\author{Haotong Xie}
\affiliation{%
  \institution{Shanghai University of Finance and Economics}
  \city{Shanghai}
  \country{China}}
\email{2023111809@stu.sufe.edu.cn}

\author{Wentao Zhang}
\affiliation{%
  \institution{Peking University}
  \city{Beijing}
  \country{China}}
\email{wentao.zhang@pku.edu.cn}
\renewcommand{\shortauthors}{Trovato et al.}

\begin{abstract}
Knowledge graph question generation~(KGQG) aims to generate natural-language questions from structured graph evidence. Existing KGQG benchmarks, however, are mostly built on static knowledge graphs and do not encode the temporal scopes of graph facts. As a result, they cannot evaluate whether generated questions faithfully preserve temporal validity, event ordering, and answer-determining temporal constraints. In this paper, we study temporal knowledge graph question generation~(TKGQG), where a generated question must be faithful to both the support subgraph and the temporal constraints required to identify the target answer. We propose \sys, the first temporally expressive and hop-bounded benchmark construction framework for TKGQG. \sys integrates a comprehensive temporal-constraint taxonomy, topology-temporal subgraph sampling, and trace-grounded question generation to construct temporally faithful questions. The framework produces four benchmark datasets from heterogeneous temporal knowledge graphs, totaling 16{,}011 verified questions. We evaluate representative LLM-based KGQG methods and prompting baselines across diverse TKGQG settings, including temporal-constraint counts, topological templates, and temporal-constraint types. The results show that existing methods struggle to preserve temporal constraints, especially under multi-constraint settings and harder temporal-constraint types. These findings reveal a clear gap between static KGQG and TKGQG, and establish \sys as a challenging testbed for temporally faithful question generation.  The code and benchmarks are available at \url{https://anonymous.4open.science/r/ChronoQG-AB85}.
\end{abstract}

\maketitle

\section{Introduction}
\label{sec:intro}
Knowledge graph question generation~(KGQG) is a core task for converting structured graph evidence into natural-language question--answer pairs. Given a support subgraph and a target answer, KGQG aims to generate a question that is answerable from the provided graph and grounded in the specified answer. Such generated questions provide scalable supervision for training and evaluating question answering systems~\cite{liu2025enhancing,guo2022dsm,liu2025fkqg}, and also support the construction of question-asking dialogue agents~\cite{zeng2020exploiting}. Early KGQG methods typically rely on graph-to-sequence architectures that encode graph structures and decode natural-language questions~\cite{fei2022lfkqg,chen2023toward,bi2025lemon,ren2025r2dqg}. More recently, large language models~(LLMs) have been adopted to serialize graph contexts into prompts and leverage instruction-following capabilities for generating more fluent and diverse questions with minimal task-specific supervision~\cite{liang2023prompting,guo2024sgsh,zhao2024zero,liu2025fkqg}.

Despite this progress, KGQG is still predominantly evaluated on static benchmarks, such as SimpleQuestion~\cite{simplequestion}, WebQuestions~\cite{berant2013semantic}, and PathQuestions~\cite{zhou2018interpretable}. These benchmarks model facts as timeless triples and omit temporal scopes, such as timestamps and validity intervals. Consequently, they cannot test whether a generation method faithfully expresses temporal validity, event ordering, fact evolution, or answer-determining temporal constraints. This limitation motivates temporal knowledge graph question generation~(TKGQG), where the input consists of a time-aware support subgraph and a target answer, and the generated question must preserve both the graph structure and the temporal constraints required to identify the answer.

\begin{figure*}[t!]
\centering
\includegraphics[scale=0.52]{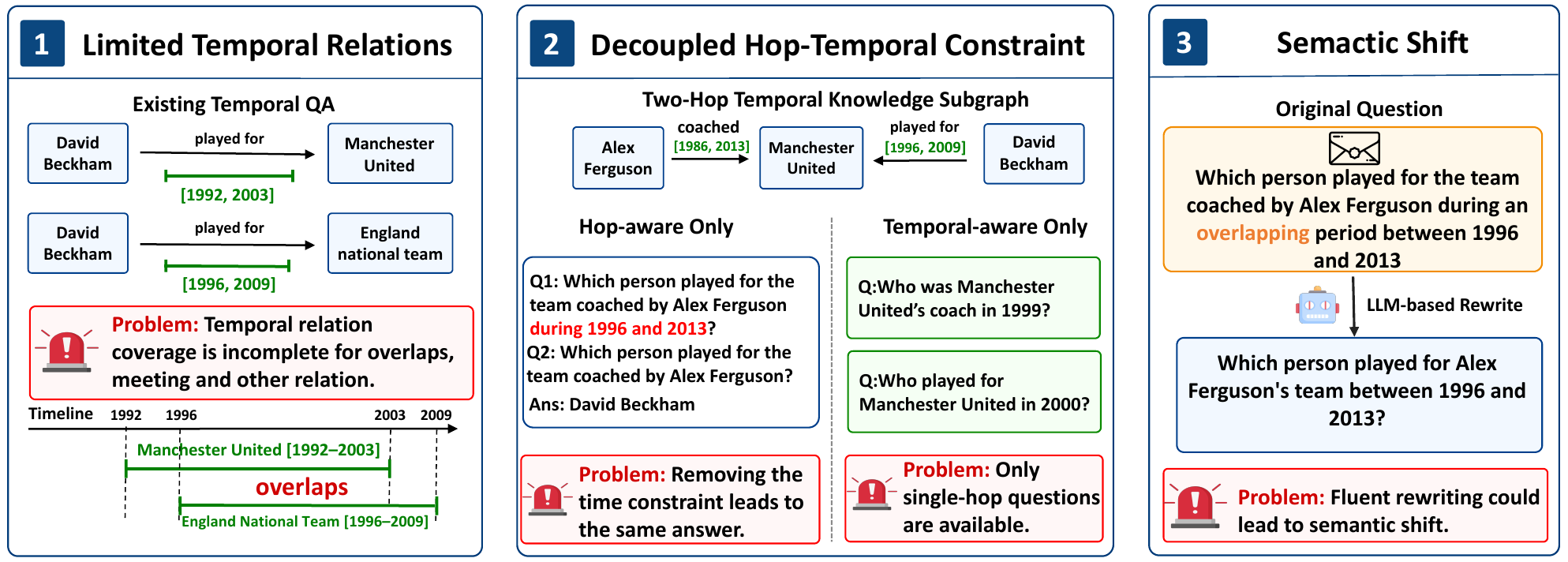}
\caption{Limitation illustration.}
\label{fig:limitations}
\end{figure*}

A natural way to construct a TKGQG benchmark is to extend a conventional KGQG pipeline: sample a subgraph and a target answer from a temporal knowledge graph~(TKG), serialize the temporal facts, and ask a model to generate a natural-language question. However, temporal benchmark construction requires more than adding timestamps to static graph evidence. As illustrated in Figure~\ref{fig:limitations}, directly adapting ordinary KGQG construction to TKGs leads to three major limitations in benchmark construction.

First, temporal constraints are incompletely modeled~\cite{saxena2021question,sun2025timelinekgqa}. Existing construction strategies typically rely on a small set of coarse temporal expressions, such as \emph{before}, \emph{after}, and \emph{during}. However, real TKG facts exhibit richer temporal interactions, including interval overlap, boundary contact, containment, and dependencies between point-valued and interval-valued facts. For example, two football-team membership facts may overlap in time, but this relation cannot be represented if a benchmark only supports coarse temporal operators. Without a comprehensive temporal-constraint space, the benchmark cannot systematically test whether generation methods preserve fine-grained temporal semantics.

Second, graph topology and temporal constraints are often not coupled. A reliable TKGQG benchmark should ensure that the target answer is identified only through the joint effect of subgraph structure and temporal constraints. However, direct construction pipelines typically sample a support subgraph first and then attach temporal expressions during question realization~\cite{chen2022temporal,zhang2024mustq}. This can produce temporal phrases that are syntactically present but semantically unnecessary: removing the temporal condition may leave the answer unchanged. Conversely, a pipeline may create multi-hop graph questions and temporal questions as separate cases, without requiring temporal filters to operate over the candidate set induced by the same hop-bounded subgraph. Such instances fail to test whether a generation method can preserve the coupling between structural reasoning and answer-determining temporal constraints in candidate space.

Third, natural-language generation may introduce semantic shift. Template-based construction can preserve topology-temporal semantics, but the resulting questions are often rigid and unnatural. LLM-based rewriting improves fluency, but may alter the intended meaning. For example, a question asking which person played for the team coached by Alex Ferguson during a specific temporal overlap may be rewritten as “Who played for Alex Ferguson’s team between 1996 and 2013?”, drifting the temporal constraint from an overlap to a simple between. Although fluent, the rewrite changes the temporal binding and weakens the intended reasoning semantics. Without verification, such shifts can silently enter the benchmark and reduce its reliability.

To address these limitations, we propose \sys, a benchmark construction framework for TKGQG. Given a TKG, \sys constructs temporally expressive and hop-bounded benchmark instances through three components. First, \sys defines a comprehensive taxonomy of temporal constraints by integrating Allen's interval algebra~\cite{allen1983maintaining} with point algebra~\cite{vilain1990constraint}. The taxonomy covers interval--interval, point--interval, interval--point, point--point, and ordinal constraints, providing a unified operator space for constructing fine-grained temporal questions. Second, \sys performs topology-temporal subgraph sampling. Rather than sampling a subgraph and attaching temporal phrases afterward, it maintains a shared candidate answer space and progressively applies structural traversals, structural filters, and temporal filters. Each accepted action must update or shrink the candidate set, ensuring that the final answer is determined by the joint effect of hop-bounded subgraph topology and temporal constraints. Third, \sys conducts trace-grounded question generation. It first constructs a template question from the accepted sampling trace, then rewrites the template into fluent natural language through verifier-guided agentic rewriting. Failed rewrites are repaired or discarded, reducing semantic shift and answer leakage during benchmark construction.

Using \sys, we construct a new TKGQG benchmark from two heterogeneous source TKGs with different temporal granularities. We instantiate \sys on CronKG~\cite{saxena2021question}, a year-level TKG, and EventKG~\cite{gottschalk2018eventkg,gottschalk2020eventkg+}, a day-level event-centric TKG. This yields four verified benchmark splits: ChronoQG-Cron-S, ChronoQG-Cron-M, ChronoQG-Event-S, and ChronoQG-Event-M, where ``-S'' denotes single-constraint questions and ``-M'' denotes multi-constraint questions. In total, the benchmark contains 16{,}011 verified questions after redundancy removal, covering diverse topological templates, temporal-constraint types, and temporal-constraint counts. We evaluate representative LLM-based KGQG methods adapted to temporal inputs, together with direct prompting baselines, on the proposed benchmark. The results show that existing static KGQG methods struggle to preserve temporal constraints, especially under multi-constraint settings and harder temporal-constraint types. Moreover, adding temporal constraints causes performance degradation comparable to increasing graph hops, indicating that temporal-constraint count is a key difficulty dimension for TKGQG. These findings reveal a clear gap between static KGQG and TKGQG, and demonstrate that \sys provides a challenging and diagnostic benchmark for temporally faithful question generation.

\section{Related Work}
\label{sec:related}
\subsection{Temporal Knowledge Graphs}
\label{sec:related-tkg}
Temporal knowledge graphs~(TKGs) extend conventional knowledge graphs~(KGs) by associating facts with temporal scopes, such as timestamps or validity intervals~\cite{liu2023retia,leetaru2013gdelt}. These temporal annotations allow TKGs to represent when facts hold, thereby supporting fact evolution modeling, temporal validity reasoning, event ordering, and other time-sensitive applications. Several large-scale temporal knowledge resources have been constructed for these purposes~\cite{leetaru2013gdelt,obrien2010crisis}. For example, Wikidata-derived TKGs such as CronKG~\cite{saxena2021question} provide year-level temporal annotations, while EventKG~\cite{gottschalk2018eventkg,gottschalk2020eventkg+} primarily focuses on a large number of day-level event-centric facts and integrates information from multiple sources, including DBpedia, YAGO, and Wikidata.

Built on these resources, many temporal question answering~(QA) datasets have been proposed to evaluate time-aware reasoning over structured knowledge. TempQuestions~\cite{jia2018tempquestions} categorizes temporal questions into explicit temporal, implicit temporal, temporal-answer, and ordinal-constraint questions. CronQuestions~\cite{saxena2021question} constructs temporal questions over a TKG and covers simple-time, simple-entity, before/after, first/last, and time-join questions. MultiTQ~\cite{chen2023multi} introduces multi-granularity temporal questions, supporting reasoning over temporal information at different granularities. MusTQ~\cite{zhang2024mustq} studies multi-step temporal reasoning and derives reasoning types from measure-theoretic operations over time. More recent datasets further incorporate relation-level temporal modeling. TimelineKGQA~\cite{sun2025timelinekgqa} uses Allen-style interval relations during temporal QA-pair generation, while TDBench~\cite{kim2026harnessing} compiles Allen relations into temporal SQL predicates to evaluate temporal reasoning over databases. These datasets provide important foundations for temporal reasoning, but they are designed primarily for answering temporal questions. In contrast, TKGQG requires generating questions from TKG evidence while preserving both the subgraph and the answer-determining temporal constraints.

\subsection{Knowledge Graph Question Generation}
\label{sec:related-kgqg}
Knowledge graph question generation~(KGQG) aims to generate natural-language questions from structured graph evidence and a target answer. Existing methods can be broadly categorized into template-based, neural, and large language model~(LLM)-based approaches. Template-based methods use hand-crafted question patterns and graph matching rules to transform KG structures into questions~\cite{seyler2017knowledge,seyler2015generating}. Neural methods learn graph-to-question mappings from annotated graph-question pairs, typically by encoding the input graph and decoding the corresponding question~\cite{liu2019generating,chen2023toward,fei2022lfkqg,guo2024diversifying,ren2025r2dqg}. A typical example is Graph2Seq~\cite{xu2018graph2seq}, an end-to-end model with an improved graph neural network and attention mechanism to map graph-structured inputs to sequences. More recently, LLM-based methods have improved fluency and diversity by linearizing graph contexts into prompts and leveraging instruction-following capabilities~\cite{zhao2024zero,guo2024sgsh,liang2023prompting,liu2025fkqg}. Representative methods include KQG-CoT+~\cite{liang2023prompting}, which prompts LLMs with chain-of-thought decomposition; SGSH~\cite{guo2024sgsh}, which uses skeleton-guided heuristics; RoleAgent~\cite{zhao2024zero}, which formulates KGQG as a multi-role editorial workflow; R2DQG~\cite{ren2025r2dqg}, which balances diversity and quality via template-guided draft generation and semantic refinement.

KGQG benchmarks have played an important role in driving method development by providing standardized graph-question pairs. Early benchmarks such as WebQuestions~\cite{berant2013semantic} and SimpleQuestion~\cite{simplequestion} focus on relatively simple graph patterns, while later datasets introduce controllable difficulty through longer paths and more complex graph structures~\cite{zhou2018interpretable}. These benchmarks support evaluation of whether KGQG methods can preserve graph structure, ground questions in the target answer, and produce fluent natural-language questions. However, they are built on static KGs, where facts are modeled as timeless triples. As a result, they cannot evaluate whether generated questions preserve temporal validity, event ordering, or answer-determining temporal constraints. This gap motivates TKGQG, where questions are generated from time-aware graph evidence and must remain faithful to both the support subgraph and the temporal constraints.

\section{Preliminaries}
\label{sec:preliminaries}
\noindent\textbf{Knowledge Graph.}
Let $\mathcal{E}$ and $\mathcal{R}$ denote the sets of entities and relations, respectively. A knowledge graph~(KG) is defined as $\mathcal{G}=(\mathcal{E},\mathcal{R},\mathcal{F})$, where $\mathcal{F}$ is a set of relational facts~\cite{zhang2025historically}. Each fact is represented as a triple $(e_s,r,e_o)$, where $e_s,e_o\in\mathcal{E}$ are the subject and object entities, and $r\in\mathcal{R}$ is the relation connecting them. Given a subset of facts, we use $\mathcal{S}^{(i)}\subseteq\mathcal{F}$ to denote a support subgraph sampled from $\mathcal{G}$. In conventional KGQG benchmarks, such facts are treated as timeless triples, and the generated question is expected to preserve the structural reasoning pattern in the support subgraph.

\noindent\textbf{Temporal Knowledge Graph.}
Temporal knowledge graphs~(TKGs) extend KGs by associating facts with temporal scopes, such as timestamps or validity intervals~\cite{zhang2025historically,gottschalk2020eventkg+,leetaru2013gdelt,chen2024local}. A temporal knowledge graph~(TKG) is defined as $\mathcal{G}_{T}=(\mathcal{E},\mathcal{R},\mathcal{T},\mathcal{F}_T)$, where $\mathcal{F}_{T}$ is a set of temporal facts~\cite{gottschalk2020eventkg+,leetaru2013gdelt,chen2024local}. Each temporal fact is represented as $(e_s,r,e_o,\tau)$, where $e_s,e_o\in\mathcal{E}$ are the subject and object entities, $r\in\mathcal{R}$ is the relation, and $\tau\in\mathcal{T}$ denotes its temporal annotation. In this work, $\tau$ can be either a point timestamp $t$ or an interval $[t_s,t_e]$ with start time $t_s$ and end time $t_e$. A point fact can be viewed as a special interval with $t_s=t_e$. Given a subset of temporal facts, we use $\mathcal{S}_T^{(i)}\subseteq\mathcal{F}_{T}$ to denote a temporal support subgraph.

\begin{figure*}[t!]
    \centering
    \includegraphics[scale=0.4]{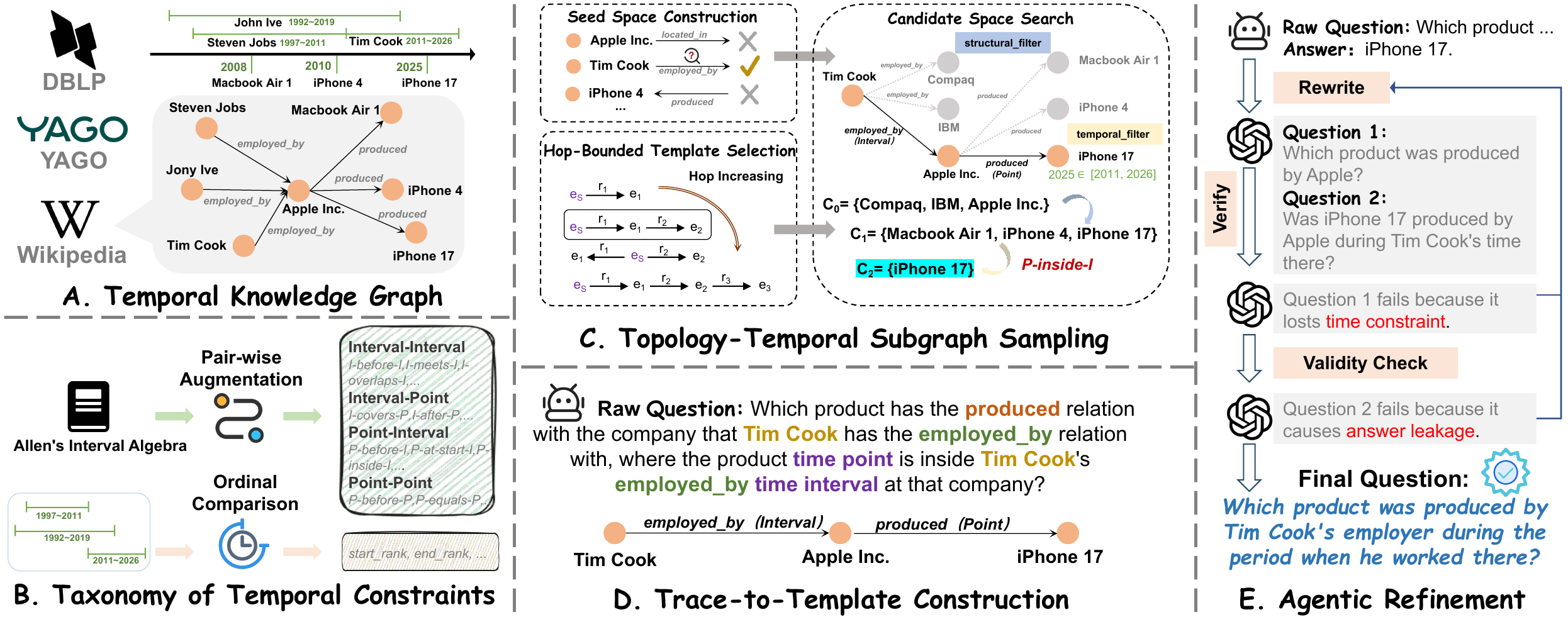}
    \caption{Overview of \sys. Given heterogeneous temporal knowledge graphs, \sys first defines a comprehensive temporal-constraint taxonomy over point- and interval-valued facts. It then performs topology-temporal subgraph sampling, where hop-bounded templates and temporal filters jointly narrow a shared candidate answer space. Finally, \sys constructs trace-grounded template questions and applies verifier-guided agentic rewriting to obtain fluent questions.}
    \label{fig:framework}
\end{figure*}

\noindent\textbf{KGQG Benchmark.}
Knowledge graph question generation~(KGQG) aims to generate a natural-language question from structured graph evidence and a target answer. A KGQG benchmark is defined as
\begin{equation}
\mathcal{D}_{KGQG}={\left\{\left(\mathcal{S}^{(i)},a_i,q_i\right)\right\}}_{i=1}^{N},
\end{equation}
where $\mathcal{S}^{(i)}$ is a support subgraph sampled from a knowledge graph $\mathcal{G}$, $a_i$ is the target answer, and $q_i$ is the reference question. A valid instance requires $q_i$ to be answerable from $\mathcal{S}^{(i)}$ with answer $a_i$, i.e., $\mathrm{Ans}(q_i,\mathcal{S}^{(i)})=a_i$. Existing KGQG benchmarks mainly focus on static graph, where facts are modeled as timeless triples and the generated question is expected to preserve the structural reasoning pattern that connects the support subgraph to the target answer.

\noindent\textbf{TKGQG Benchmark.}
Temporal knowledge graph question generation~(TKGQG) extends KGQG from static graph evidence to time-aware graph evidence. A TKGQG benchmark is defined as
\begin{equation}
\mathcal{D}_{TKGQG}=\left\{\left(\mathcal{S}^{(i)}_T,a_i,\Gamma_i,q_i\right)\right\}_{i=1}^{N},
\end{equation}
where $\mathcal{S}^{(i)}_T$ is a temporal support subgraph, $a_i$ is the target answer, $\Gamma_i$ is the set of temporal constraints that determine the answer, and $q_i$ is the corresponding natural-language question. A valid TKGQG instance must satisfy two requirements. First, the question should be structurally grounded: it must reflect the graph reasoning structure in $\mathcal{S}^{(i)}_T$. Second, it should be temporally faithful: it must preserve the temporal constraints in $\Gamma_i$ such that $\mathrm{Ans}(q_i,\mathcal{S}^{(i)}_T,\Gamma_i)=a_i$. Therefore, a TKGQG benchmark should evaluate not only whether a generated question is fluent and graph-grounded, but also whether it faithfully expresses the temporal conditions.

\section{Benchmark Design}
\label{sec:framework}
In this section, we introduce \sys, a benchmark construction framework for TKGQG. As discussed in Sec.~\ref{sec:intro}, directly adapting conventional KGQG construction pipelines to TKGs is insufficient~\cite{saxena2021question,gottschalk2018eventkg,gottschalk2020eventkg+}. A reliable TKGQG benchmark must support fine-grained temporal constraints, couple hop-bounded topological reasoning with temporal filtering, and preserve graph-temporal semantics during question generation. To this end, \sys consists of three components. First, it defines a comprehensive taxonomy of temporal constraints over point- and interval-valued facts, providing the operator space for temporally fine-grained benchmark construction. Second, it performs topology-temporal subgraph sampling, where hop-bounded templates and temporal operators are applied over a shared candidate answer space so that the final answer is jointly determined by subgraph topology and temporal constraints. Third, it conducts trace-grounded question generation: \sys first constructs a template question from the accepted sampling trace and then rewrites it into fluent natural language through verifier-guided agentic rewriting. Figure~\ref{fig:framework} summarizes the overall construction process.

\begin{table*}[t!]
\centering
\caption{Temporal taxonomy used by \sys.}
\small
\setlength{\tabcolsep}{5.5pt}
\renewcommand{\arraystretch}{1.05}
\begin{tabular}{lcc}
\toprule
\textbf{Family} & \textbf{Number} & \textbf{Operators} \\
\midrule
Interval--Interval (I--I)
& 13
& \makecell[l]{%
\emph{I-before-I}, \emph{I-meets-I}, \emph{I-overlaps-I}, \emph{I-finished-by-I}, \emph{I-contains-I}, \emph{I-starts-I}, \emph{I-equals-I}, \\
\emph{I-started-by-I}, \emph{I-during-I}, \emph{I-finishes-I}, \emph{I-overlapped-by-I}, \emph{I-met-by-I}, and \emph{I-after-I}
} \\
\midrule
Point--Interval (P--I)
& 5
& \makecell[l]{%
\emph{P-before-I}, \emph{P-at-start-I}, \emph{P-inside-I}, \emph{P-at-end-I}, and \emph{P-after-I}
} \\
\midrule
Interval--Point (I--P)
& 5
& \makecell[l]{%
\emph{I-before-P}, \emph{I-ends-at-P}, \emph{I-covers-P}, \emph{I-starts-at-P}, and \emph{I-after-P}
} \\
\midrule
Point--Point (P--P)
& 3
& \makecell[l]{%
\emph{P-before-P}, \emph{P-equals-P}, and \emph{P-after-P}
} \\
\midrule
Ordinal
& 3
& \makecell[l]{%
\textit{start\_rank}, \textit{end\_rank}, and \textit{duration\_rank}
} \\
\bottomrule
\end{tabular}
\label{tab:taxonomy-summary}
\end{table*}

\subsection{A Comprehensive Taxonomy of Temporal Constraints}
\label{sec:taxonomy}
A TKGQG benchmark requires a temporal operator space that can uniformly characterize temporal dependencies among heterogeneous facts. In real-world TKGs, facts may be annotated with either point timestamps, such as birth dates, or interval timestamps, such as office terms and memberships. As a result, coarse temporal operators such as \emph{before}, \emph{after}, and \emph{during} are insufficient for generating temporally fine-grained questions~\cite{saxena2021question,liu2024towards}.

\sys therefore introduces a comprehensive taxonomy of temporal constraints by integrating Allen's interval algebra~\cite{allen1983maintaining} with point algebra~\cite{vilain1990constraint}. Allen's interval algebra defines 13 mutually exclusive and collectively exhaustive relations between two intervals: \emph{before}, \emph{meets}, \emph{overlaps}, \emph{finished-by}, \emph{contains}, \emph{starts}, \emph{equals}, \emph{started-by}, \emph{during}, \emph{finishes}, \emph{overlapped-by}, \emph{met-by}, and \emph{after}. However, Allen's algebra only models interval--interval relations. To support point-valued facts, we further incorporate point algebra, which defines three primitive relations between time points: \emph{before}, \emph{equal}, and \emph{after}. Based on the temporal type of each fact, each pairwise temporal constraint in \sys falls into one of the following four families:

\begin{itemize}
\item \textbf{Interval--Interval (I--I).} This family follows Allen's interval algebra and models temporal constraints between two interval-valued facts. It contains the 13 interval relations: \emph{I-before-I}, \emph{I-meets-I}, \emph{I-overlaps-I}, \emph{I-finished-by-I}, \emph{I-contains-I}, \emph{I-starts-I}, \emph{I-equals-I}, \emph{I-started-by-I}, \emph{I-during-I}, \emph{I-finishes-I}, \emph{I-overlapped-by-I}, \emph{I-met-by-I}, and \emph{I-after-I}.

\item \textbf{Point--Interval (P--I).} This family specifies the position of a point-valued fact with respect to an interval-valued fact. It includes \emph{P-before-I}, \emph{P-at-start-I}, \emph{P-inside-I}, \emph{P-at-end-I}, and \emph{P-after-I}.

\item \textbf{Interval--Point (I--P).} This family specifies the position of an interval-valued fact with respect to a point-valued fact. It includes \emph{I-before-P}, \emph{I-ends-at-P}, \emph{I-covers-P}, \emph{I-starts-at-P}, and \emph{I-after-P}.

\item \textbf{Point--Point (P--P).} This family models ordering and equality constraints between two point-valued facts. It includes \emph{P-before-P}, \emph{P-equals-P}, and \emph{P-after-P}.
\end{itemize}

The four families above define pairwise temporal constraints between two facts. However, benchmark instances in TKGQG may also involve temporal comparisons over a set of candidate facts, rather than a single pair. Pairwise constraints cannot express questions that require identifying the earliest, latest, or $k$-th fact satisfying a temporal condition. To support such cases, \sys further introduces three ordinal operators: \textit{start\_rank}, \textit{end\_rank}, and \textit{duration\_rank}. These operators rank a candidate set by start time, end time, or duration, and then select the candidate at the required rank. Overall, the taxonomy contains 26 pairwise temporal relations and 3 ordinal operators. This taxonomy defines the temporal operator space used by \sys for constraint sampling and benchmark construction. Table~\ref{tab:taxonomy-summary} summarizes the operator families.

\subsection{Topology-Temporal Subgraph Sampling}
\label{sec:sampling}
Given the temporal-constraint taxonomy, \sys constructs benchmark instances through topology-temporal subgraph sampling. The goal is to generate reasoning traces in which subgraph topology and temporal constraints jointly determine the target answer. Instead of first sampling a subgraph and then attaching temporal expressions afterward~\cite{chen2022temporal}, \sys maintains a shared candidate answer space throughout the sampling process. Each structural traversal or temporal condition is accepted only when it contributes to narrowing this space. The procedure consists of three stages: seed space construction, hop-bounded template selection, and candidate space search.

\subsubsection{Seed Space Construction}
\sys first constructs a temporally valid seed space from the source TKG. Each relation is assigned one of three temporal types: \emph{Point}, \emph{Interval}, or \emph{Deleted}. Relations marked as \emph{Deleted} are either semantically ambiguous or associated with facts that mix point and interval timestamps, and are therefore excluded from sampling. \sys also removes invalid temporal facts, including facts with inconsistent timestamps where $t_s > t_e$ for the interval $[t_s, t_e]$ and facts attached to deleted relations. This step ensures that subsequent temporal constraints are applied only to facts with well-defined temporal semantics.

After temporal annotation, \sys samples a root seed from the remaining temporally valid facts. A root seed consists of an anchor entity and a directed relation, and is used to retrieve the initial candidate answer set. For example, given the seed $(\emph{educated\_at}, \emph{University of Calgary})$ in the inverse direction, \sys retrieves all entities whose \emph{educated\_at} relation points to \emph{University of Calgary}. Formally, this seed defines the initial candidate set as
\begin{equation}\small
C_0=\left\{e \mid (e,\emph{educated\_at},\emph{University of Calgary},\tau_e)\in \mathcal{F}_{T}\right\},
\end{equation}
where $\tau_e\in\mathcal{T}$ denotes the timestamp associated with the fact. \sys keeps only seeds satisfying $|C_0|\in[5,50]$. This range provides enough candidates for composing structural and temporal constraints, while avoiding overly broad candidate spaces that make controlled instance construction difficult.

\subsubsection{Hop-Bounded Template Selection}
After selecting a root seed, \sys chooses a topological template that specifies how the reasoning structure expands from the root. A template abstracts the structure of a sampled reasoning subgraph: each node denotes an entity variable, and each directed edge denotes a relation. \sys considers the following representative hop-bounded templates:
\begin{equation}\small
e_a\to e_b,\quad
e_c\leftarrow e_a\to e_b,\quad
e_a\to e_b\to e_c,\quad
e_a\to e_b\to e_c\to e_d.
\end{equation}
Here, each $e_i$ denotes an entity variable, and $e_i\to e_j$ means that there exists a relation from $e_i$ to $e_j$ in the TKG. The template $e_a\to e_b$ represents one-hop reasoning from the root to candidate answers. The template $e_c\leftarrow e_a\to e_b$ represents a branching structure, where candidates are further constrained by another relation incident to the root. The templates $e_a\to e_b\to e_c$ and $e_a\to e_b\to e_c\to e_d$ capture two-hop and three-hop chain reasoning respectively. \sys does not use longer templates because temporal filtering already increases reasoning difficulty, and long relational chains substantially reduce answerability after candidate filtering. For fine-grained TKGs such as EventKG~\cite{gottschalk2018eventkg}, longer templates are harder to instantiate after candidate filtering. \sys therefore allocates a larger sampling budget to multi-hop templates to maintain sufficient coverage of complex reasoning cases.

\subsubsection{Candidate Space Search}
Starting from $C_0$, \sys expands the selected template while progressively updating the candidate answer space. Let $C_t$ denote the candidate set after the $t$-th sampling step. At each step, the sampler applies one action to either extend the sampled structure or impose a temporal condition, producing a new candidate set $C_{t+1}$. The process stops once the candidate set contains only a single entity, i.e., $|C_{t+1}|=1$, which is used as the target answer. 

The sampler supports three types of candidate-space updating actions. First, \texttt{forward\_transition} follows a relation traversal specified by the selected template. For a chain template such as $e_a\to e_b\to e_c$, after candidates for $e_b$ are obtained, the sampler follows the next relation and replaces each candidate $e_b$ with the reachable entities that can instantiate $e_c$. This action extends the reasoning trace along the selected structure while updating the candidate answer set. Second, \texttt{structural\_filter} adds an additional edge constraint and removes candidates that do not satisfy it. For example, consider the branching template $e_c\leftarrow e_a\to e_b$. If the root seed fixes $e_b$ as \emph{University of Calgary} and retrieves candidates for $e_a$ through the inverse direction of \emph{educated\_at}, then $C_0$ contains people educated at \emph{University of Calgary}. To instantiate the other branch, the sampler may fix $e_c$ as \emph{University of Alberta} and require each candidate $e_a$ to also satisfy the edge $e_a\to e_c$ through \emph{educated\_at}. The candidate set is then reduced to people who attended both universities. Third, \texttt{temporal\_filter} imposes a temporal constraint on facts already associated with the sampled structure. Under the same branching template, the sampler compares the timestamp of $(e_a,\emph{educated\_at},e_b,\tau_b)$ with that of $(e_a,\emph{educated\_at},e_c,\tau_c)$. According to the temporal types of $\tau_b$ and $\tau_c$, it selects a compatible operator from the taxonomy and retains only candidates whose timestamps satisfy the constraint. For example, the sampler may require the \emph{University of Calgary} education period to start before and overlap with the \emph{University of Alberta} education period.

The progressive construction above enforces constraint necessity during sampling. In this way, \sys filters out the pseudo-temporal conditions~\cite{chen2022temporal}, namely those whose removal leaves the correct answers unchanged. This prevents temporal expressions from being added as decorative constraints to a pre-sampled subgraph, and ensures that they are genuinely involved in determining the final answer. For instances with multiple temporal constraints, \sys further checks whether the constraints are mutually non-redundant. For example, ``after 2000'' becomes unnecessary once ``between 2005 and 2010'' is imposed, because the latter already implies the former. \sys detects such cases by comparing the candidate sets induced by temporal constraints over the same original candidate pool. If one constraint yields a proper subset of another, the weaker constraint is removed. Instances that no longer satisfy the required constraint setting after this removal are discarded. 


\subsection{Trace-Grounded Question Generation}
\label{sec:postproc}
After topology-temporal subgraph sampling, each instance is represented by a reasoning trace, which records how the initial candidate answer set is progressively narrowed by structural traversals, structural filters, and temporal constraints. \sys converts this trace into a benchmark question in two steps. It first constructs a template question that preserves the sampled semantics, and then rewrites it into a natural question under verifier guidance.

\subsubsection{Template Construction from Sampling Traces}
\sys first serializes each accepted sampling trajectory into an intermediate benchmark record. Each record stores the topological template, hop count, temporal-constraint count, ordered sampling steps, temporal constraints, support facts, target answer, and a template question. The ordered sampling steps describe how the candidate answer set is reduced during sampling. The template question deterministically verbalizes the accepted structural and temporal constraints.

\begin{figure}
    \centering
    \includegraphics[scale=0.3]{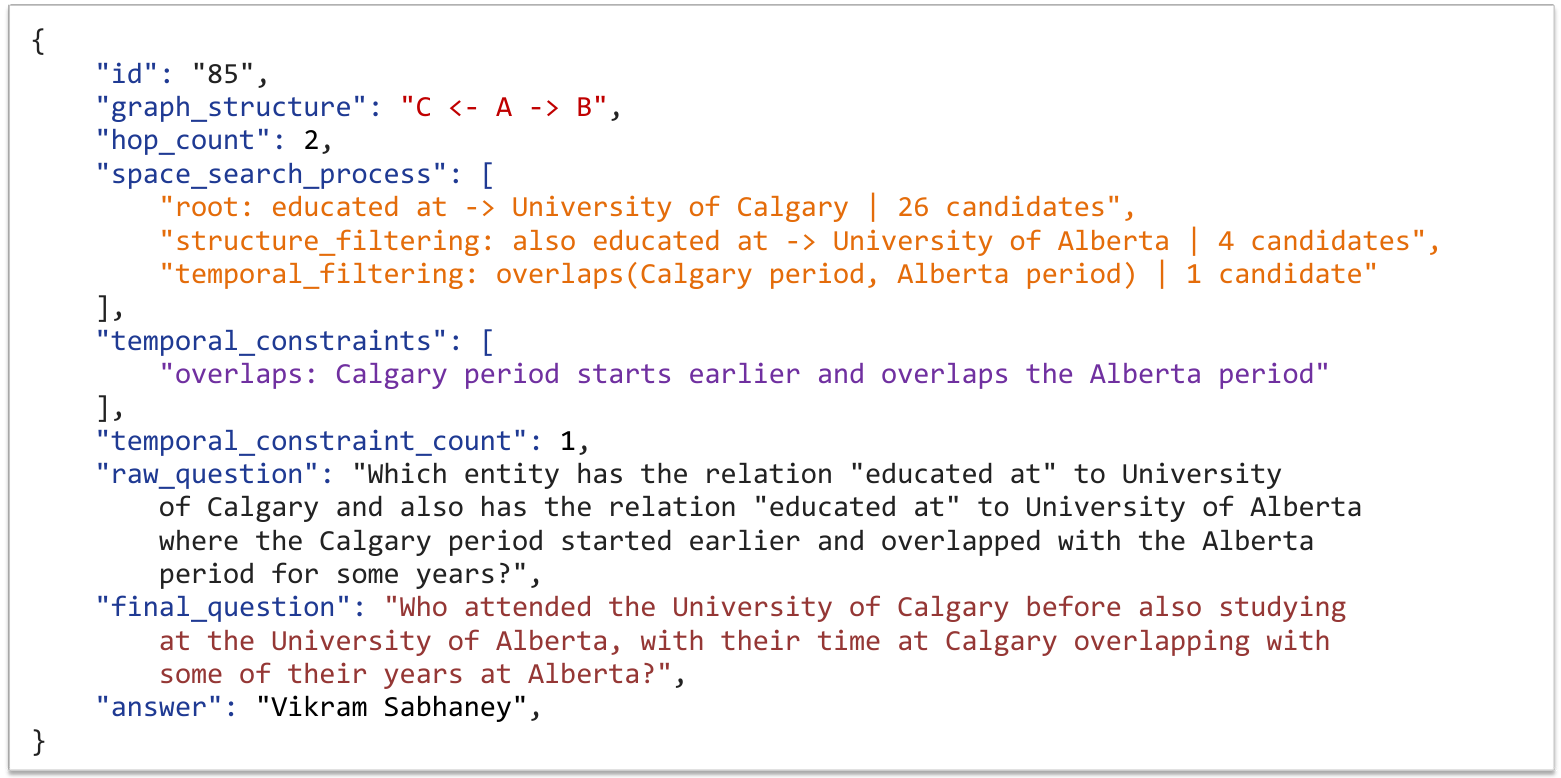}
    \caption{Instance of sampling record.}
    \label{fig:benchmark-case}
\end{figure}

Figure~\ref{fig:benchmark-case} clearly illustrates this process with a concrete instance (\emph{id: 85}). The instance follows the branching template $e_c\leftarrow e_a\to e_b$. The root seed fixes $e_b$ as \emph{University of Calgary} and retrieves candidate entities $e_a$ whose \emph{educated\_at} relation points to this university, yielding an initial candidate set of 26 entities. The sampler then instantiates the other branch by fixing $e_c$ as \emph{University of Alberta} and requiring the same candidates to also satisfy the relation $e_a\to e_c$ through \emph{educated\_at}. This structural filter reduces the candidate set from 26 to 4. Since multiple candidates still satisfy the sampled topology, the sampler further applies a strict temporal constraint between the two education facts. Specifically, it requires the Calgary education period to start earlier and overlap with the Alberta education period. This temporal filter reduces the candidate set to a unique answer, \emph{Vikram Sabhaney}.

\begin{figure}[t!]
    \centering
    \subfigure[Cron-S]{
    \begin{minipage}{0.45\linewidth}
        \centering
        \includegraphics[scale=0.26]{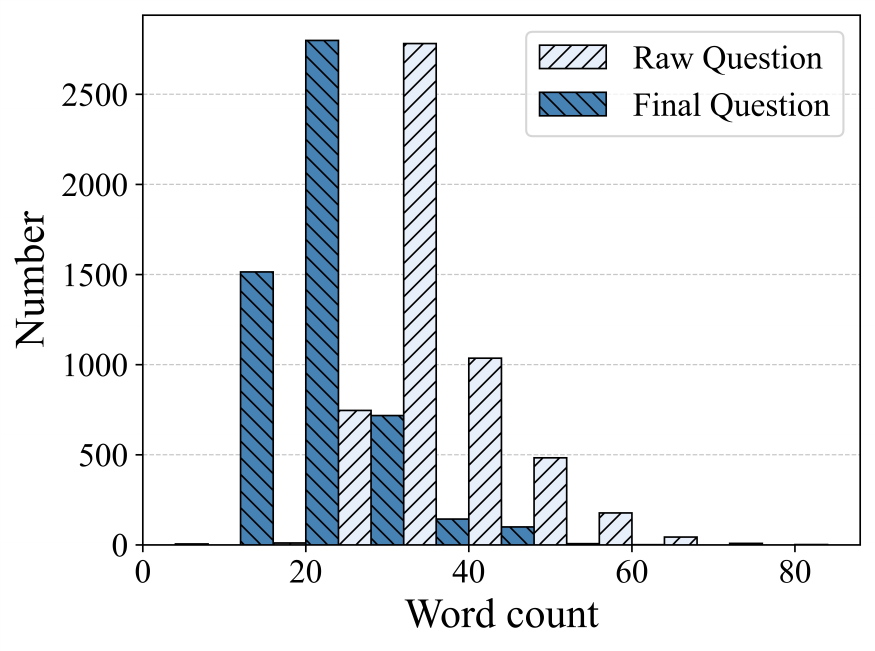}
    \end{minipage}}
    	\subfigure[Cron-M]{
    \begin{minipage}{0.45\linewidth}
        \centering
        \includegraphics[scale=0.26]{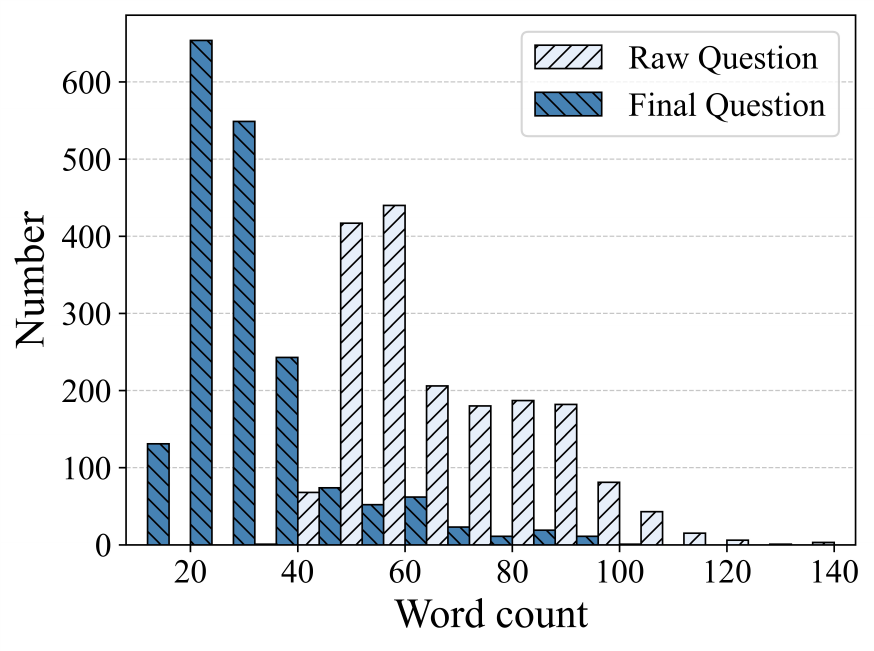}
    \end{minipage}}

	\subfigure[Event-S]{
    \begin{minipage}{0.45\linewidth}
        \centering
        \includegraphics[scale=0.26]{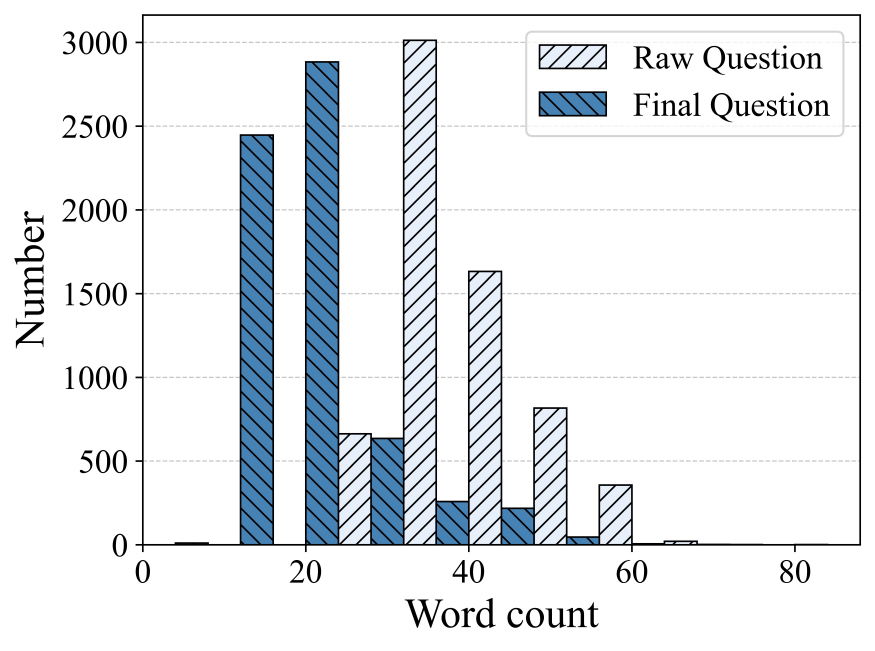}
    \end{minipage}}
    	\subfigure[Event-M]{
    \begin{minipage}{0.45\linewidth}
        \centering
        \includegraphics[scale=0.26]{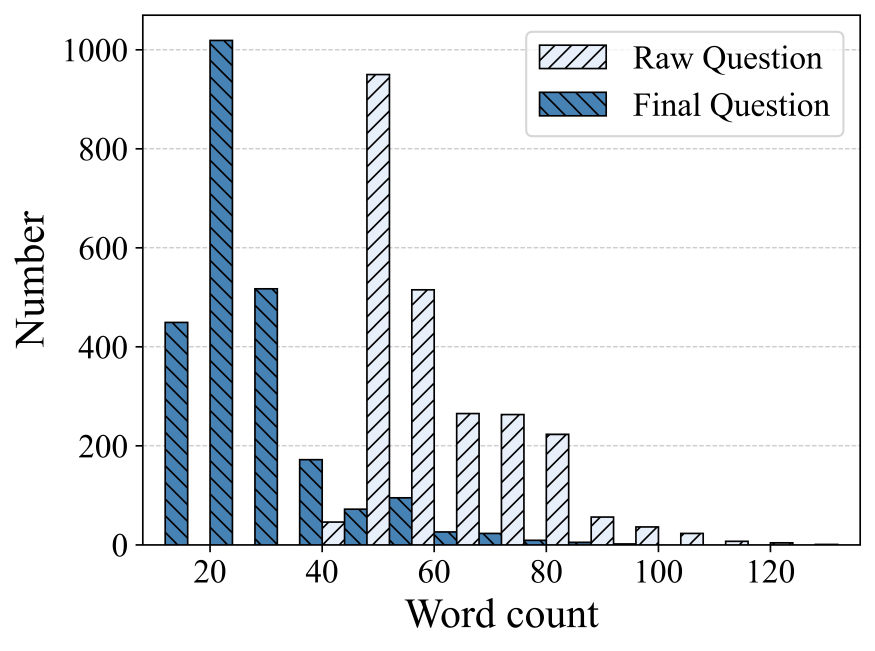}
    \end{minipage}}
    \caption{Word count distributions of raw and final questions across four benchmark datasets.}
    \label{fig:word-count-distribution}
    \vspace{-1em}
\end{figure}

Based on this trace, \sys constructs a template question by concatenating the corresponding constraint fragments. In this example, the template asks which entity has the \emph{educated\_at} relation to both \emph{University of Calgary} and \emph{University of Alberta}, where the Calgary education period starts earlier and overlaps with the Alberta education period. The template is semantically faithful because every fragment is derived from an accepted sampling step and grounded in the support facts. However, such templates are often rigid and verbose, since they directly expose KG-style relations and concatenate constraint descriptions~\cite{zhang2024mustq,chen2023multi,meem2024pat}. \sys therefore uses the template as a semantic scaffold for later rewriting, rather than directly admitting it into the benchmark.

\subsubsection{Verifier-Guided Agentic Rewriting}
Given a template question, \sys invokes an LLM rewriter to produce a fluent natural-language question. The rewriting prompt requires the model to preserve all structural and temporal constraints. It also explicitly forbids reversing temporal directions, dropping constraints, revealing the answer, or copying KG-style relation labels verbatim. We use {GPT-4o-mini}~\cite{openai2024gpt4omini} as the default rewriter. As shown in Figure~\ref{fig:word-count-distribution}, this step substantially shortens the questions and removes redundancy introduced by literal template concatenation.

The rewritten question is then checked by an independent verifier. The verifier receives the support subgraph, the target answer, and a structure hint derived from the reasoning trace. It answers the rewritten question using only the provided evidence. \sys compares the verifier prediction with the gold answer using fuzzy matching. If the predicted answer is equivalent to the gold answer, the rewrite is regarded as semantically faithful. Since the verifier directly affects benchmark quality, we select it empirically. Specifically, we evaluate four candidate models on 500 stratified template questions, whose semantics are reliable by construction, and choose the model with the best cost--accuracy tradeoff. We observe in Table~\ref{tab:verifier-accuracy} that {GPT-5-mini}~\cite{singh2025openai} achieves 92.0\% overall accuracy and use it as the default verifier. For the samples that failed this round, \sys have arranged for the error analysis agent to analyze the causes of the failure or to determine whether the verifier mistakenly killed the correct samples. The second rewriter is given the original template question, the failed rewrite, the verifier's predicted answer, the failure reason, and the support facts. It is then asked to generate a corrected rewrite. The repaired question is verified again under the same procedure. If it passes verification, it is admitted into the benchmark; otherwise, the instance is discarded.

Before admitting a verified question, \sys applies an answer-leakage filter as a final safeguard. A verified rewrite may still be unsuitable if it directly mentions the target answer. For each accepted question, \sys scans the question for the gold answer string and its surface aliases appearing in the support subgraph. If an answer mention appears in the question, the question is rejected and sent to one additional rewrite attempt that explicitly forbids answer leakage. If the repaired version still leaks the answer, the instance is discarded. This filter ensures that verification success comes from reasoning over topology-temporal evidence rather than from trivial answer extraction.

Finally, \sys categorizes verified questions according to their temporal-constraint count. Instances with a single temporal constraint are treated as single-constraint cases, while those with multiple temporal constraints are treated as multi-constraint cases. This categorization supports separate evaluation of different temporal reasoning complexities.

\begin{table}[t!]
\centering
\footnotesize 
\caption{Verifier accuracy (\%) stratified by relation category. I and P denote interval and point temporal expressions, respectively. Each row group contains 500 samples.
}
\label{tab:verifier-accuracy}
\begin{tabular}{lcccc}
\toprule
\textbf{Category}
  & \textbf{GPT-4o-mini} & \textbf{GPT-4o} & \textbf{GPT-5-mini} & \textbf{GPT-5} \\
\midrule
I--I & 52.5 & 75.4 & 88.7 & 90.1 \\
P--I & 60.0 & 67.5 & 100.0 & 100.0 \\
I--P & 76.0 & 84.0 & 88.0 & 100.0 \\
P--P & 66.9 & 78.8 & 96.7 & 98.0 \\
\midrule
\textbf{Overall}   & \textbf{58.6} & \textbf{76.2} & \textbf{92.0} & \textbf{93.8} \\
\bottomrule
\end{tabular}
\end{table}

\section{Experiments}
\label{sec:exp}

\subsection{Experimental Setup}
\label{sec:setup}
\subsubsection{Dataset}
We use two source TKGs with different temporal granularities and schema characteristics. \textbf{CronKG}~\cite{saxena2021question} is a Wikidata-derived TKG with year-level temporal annotations and 202 relations. \textbf{EventKG}~\cite{gottschalk2018eventkg,gottschalk2020eventkg+} is a day-level, event-centric multilingual KG with 662 relations, integrating events from large-scale knowledge graphs such as DBpedia, YAGO, and Wikidata. From these two sources, \sys constructs four benchmark splits: ChronoQG-Cron-S, ChronoQG-Cron-M, ChronoQG-Event-S, and ChronoQG-Event-M, where ``-S'' denotes single-constraint instances and ``-M'' denotes multi-constraint instances. The two source TKGs differ in temporal granularity, relation inventory, and schema design, allowing us to evaluate whether \sys can construct TKGQG benchmarks across heterogeneous temporal knowledge sources. Table~\ref{tab:datasets} reports the statistics of the four splits.

\begin{table*}[t!]
\centering\small
\caption{Statistics of the proposed benchmark datasets.}
\label{tab:datasets}
\renewcommand{\arraystretch}{1.08}
\setlength{\tabcolsep}{3.0pt}
\begin{tabular}{l|c|cccc|cccc|c}
\toprule
\textbf{Dataset}  & \textbf{Source} & \textbf{\#Entity}  & \textbf{\#Relation}  & \textbf{Avg. Hops}  & \textbf{Avg. Temporal Constraint} & \textbf{\#I-I} & \textbf{\#P-I}  & \textbf{\#I-P} & \textbf{\#P-P}  & \textbf{Total} \\  
\midrule
Cron-S  & CronKG & 34{,}306 & 88 & 1.50 & 1.00 & 2{,}924 & 433 & 412 & 1{,}444 & 5{,}213 \\
Cron-M  & CronKG & 26{,}929 & 72 & 1.46 & 2.64 & 1{,}780 & 213 & 41 & 1{,}162 & 3{,}196\\
\midrule
Event-S & EventKG  & 48{,}641 & 84 & 1.49 & 1.00 & 3{,}006 & 164 & 1{,}561 & 1{,}460 & 6{,}191 \\
Event-M & EventKG  & 24{,}504 & 80 & 1.53 & 2.34 & 1{,}051 & 53 & 1{,}039 & 1{,}247 & 3{,}390\\
\bottomrule
\end{tabular}
\end{table*}

\begin{table*}[t!]
\centering
\caption{Main results on the four \sys-generated benchmark splits. Best results are boldfaced. Cron and Event denote the source TKGs, and S/M denote single-constraint and multi-constraint splits.}
\label{tab:main-results}
\renewcommand{\arraystretch}{1.08}
\setlength{\tabcolsep}{3.0pt}
\resizebox{\textwidth}{!}{
\begin{tabular}{ll|ccc|ccc|ccc|ccc}
\toprule
\multirow{2}{*}{\bf Base LLM}
& \multirow{2}{*}{\bf Method}
& \multicolumn{3}{c|}{\bf Cron-S}
& \multicolumn{3}{c|}{\bf Cron-M}
& \multicolumn{3}{c|}{\bf Event-S}
& \multicolumn{3}{c}{\bf Event-M} \\
\cmidrule(lr){3-5} \cmidrule(lr){6-8} \cmidrule(lr){9-11} \cmidrule(lr){12-14}
& & BLEU-4 & ROUGE-2 & ROUGE-L
& BLEU-4 & ROUGE-2 & ROUGE-L
& BLEU-4 & ROUGE-2 & ROUGE-L
& BLEU-4 & ROUGE-2 & ROUGE-L \\
\midrule

\rowcolor{sectiongray}
\multicolumn{14}{c}{\textbf{\em Close-source LLMs}} \\

\multirow{3}{*}{GPT-4o-mini}
& Zero-shot
& 0.458 & 0.577 & 0.666
& 0.288 & 0.404 & 0.502
& 0.410 & 0.541 & 0.631
& 0.328 & 0.456 & 0.545 \\
& Zero-shot CoT
& 0.373 & 0.499 & 0.591
& 0.254 & 0.376 & 0.472
& 0.334 & 0.476 & 0.569
& 0.269 & 0.412 & 0.501 \\
& Few-shot
& 0.368 & 0.504 & 0.587
& 0.231 & 0.353 & 0.443
& 0.375 & 0.520 & 0.602
& 0.282 & 0.427 & 0.502 \\

\midrule

\multirow{3}{*}{GPT-4o}
& Zero-shot
& 0.398 & 0.527 & 0.631
& 0.245 & 0.371 & 0.471
& 0.348 & 0.496 & 0.612
& 0.274 & 0.417 & 0.521 \\
& Zero-shot CoT
& 0.424 & 0.550 & 0.650
& 0.267 & 0.397 & 0.490
& 0.346 & 0.485 & 0.598
& 0.273 & 0.419 & 0.520 \\
& Few-shot
& 0.404 & 0.541 & 0.628
& 0.260 & 0.383 & 0.472
& 0.376 & 0.520 & 0.602
& 0.284 & 0.429 & 0.516 \\

\midrule

\multirow{3}{*}{\small Gemini-3.1-pro-preview}
& Zero-shot
& 0.368 & 0.494 & 0.620
& 0.271 & 0.388 & 0.517
& 0.366 & 0.500 & 0.611
& 0.204 & 0.337 & 0.459 \\
& Zero-shot CoT
& 0.435 & 0.550 & 0.660
& 0.318 & 0.440 & 0.568
& 0.359 & 0.489 & 0.600
& 0.237 & 0.362 & 0.489 \\
& Few-shot
& 0.369 & 0.485 & 0.602
& 0.294 & 0.414 & 0.547
& 0.338 & 0.487 & 0.589
& 0.217 & 0.349 & 0.473 \\

\rowcolor{sectiongray}
\multicolumn{14}{c}{\textbf{\em Open-source LLMs}} \\

\multirow{3}{*}{DeepSeek-V3}
& Zero-shot
& 0.292 & 0.425 & 0.525
& 0.179 & 0.294 & 0.390
& 0.271 & 0.419 & 0.526
& 0.216 & 0.346 & 0.444 \\
& Zero-shot CoT
& 0.350 & 0.480 & 0.579
& 0.219 & 0.342 & 0.435
& 0.293 & 0.442 & 0.546
& 0.228 & 0.366 & 0.467 \\
& Few-shot
& 0.363 & 0.492 & 0.587
& 0.208 & 0.330 & 0.428
& 0.360 & 0.505 & 0.583
& 0.253 & 0.396 & 0.479 \\

\midrule

\multirow{3}{*}{GLM-4.6}
& Zero-shot
& 0.266 & 0.364 & 0.496
& 0.197 & 0.290 & 0.421
& 0.236 & 0.344 & 0.473
& 0.152 & 0.248 & 0.377\\
& Zero-shot CoT
& 0.292 & 0.400 & 0.533
& 0.211 & 0.309 & 0.453
& 0.239 & 0.350 & 0.480
& 0.188 & 0.304 & 0.443\\
& Few-shot
& 0.339 & 0.432 & 0.547
& 0.280 & 0.374 & 0.496
& 0.238 & 0.347 & 0.477
& 0.216 & 0.315 & 0.444\\

\midrule

\multirow{3}{*}{Qwen2.5-32B}
& Zero-shot
& 0.405 & 0.534 & 0.635
& 0.238 & 0.367 & 0.466
& 0.340 & 0.485 & 0.593
& 0.261 & 0.402 & 0.505 \\
& Zero-shot CoT
& 0.419 & 0.545 & 0.646
& 0.257 & 0.379 & 0.475
& 0.348 & 0.488 & 0.601
& 0.268 & 0.406 & 0.508 \\
& Few-shot
& 0.401 & 0.534 & 0.623
& 0.242 & 0.364 & 0.456
& 0.361 & 0.510 & 0.586
& 0.265 & 0.416 & 0.490 \\

\midrule
\rowcolor{sectiongray}
\multicolumn{14}{c}{\textbf{\em Full Training KGQG Methods}} \\
-- & G2S-AE
&0.494	&0.569	&0.685	
&0.335	&0.384	&0.516	
&0.424	&0.523	&0.635	
&0.344	&0.404	&0.523  \\
-- & G2S-AE-RL
&0.523	&0.606	&0.717	
&0.351	&0.399	&0.526	
&0.467	&0.565	&0.675	
&0.391	&0.470	&0.584 \\

\midrule
\rowcolor{sectiongray}
\multicolumn{14}{c}{\textbf{\em LLM-based KGQG Methods}} \\

\multirow{5}{*}{GPT-4o-mini}
& KQG-CoT
& 0.417 & 0.544 & 0.637
& 0.262 & 0.396 & 0.486
& 0.249 & 0.396 & 0.502
& 0.188 & 0.334 & 0.449 \\
& SGSH
& 0.346 & 0.453 & 0.552
& 0.238 & 0.349 & 0.439
& 0.263 & 0.403 & 0.508
& 0.306 & 0.425 & 0.520 \\
& RoleAgentQG
& 0.264 & 0.388 & 0.509
& 0.184 & 0.295 & 0.410
& 0.241 & 0.383 & 0.508
& 0.169 & 0.300 & 0.418 \\
& R2DQG 
&0.199 &0.293 &0.426
&0.167 &0.266 &0.385
&0.138 &0.237 &0.370
&0.128 &0.226 &0.348 \\
\bottomrule
\end{tabular}
}
\end{table*}

\subsubsection{Evaluation Metrics}
Consistent with previous studies on static KGQG, we evaluate generated questions using BLEU-4~\cite{papineni2002bleu}, ROUGE-2, and ROUGE-L~\cite{lin2004rouge}. BLEU measures $n$-gram precision between a generated question and the reference question, while ROUGE captures recall-oriented lexical overlap based on longest common subsequences and bigram matching. These metrics provide a standard automatic evaluation protocol for comparing different baselines.

\subsubsection{Baselines}
We evaluate three groups of baselines. The first group consists of prompting baselines designed for TKGQG. Given the serialized temporal subgraph, target answer, reasoning path, and temporal constraints, zero-shot prompting asks the model to generate the question in a single step. Zero-shot CoT first instructs the model to decompose the topological and temporal requirements, and then merge them into a natural-language question. Few-shot prompting further provides demonstrations retrieved by matching the temporal-constraint type from our taxonomy. We evaluate these prompting strategies with both closed-source LLMs, including GPT-4o-mini~\cite{openai2024gpt4omini}, GPT-4o~\cite{openai2024gpt4osystemcard}, and Gemini-3.1-pro-preview~\cite{google2026gemini31pro}, and open-source LLMs, including DeepSeek-V3~\cite{liu2024deepseek}, GLM-4.6~\cite{zhipu2025glm46}, and Qwen2.5-32B~\cite{qwen2025qwen25technicalreport}. In all prompting baselines, the reasoning path is included to make the support topology explicit. The second group includes full-training KGQG methods G2S-AE and G2S-AE-RL~\cite{xu2018graph2seq}. To adapt them to TKGQG, we augment the serialized graph input with temporal annotations and temporal-constraint information. The third group adapts representative LLM-based static KGQG methods to temporal inputs. KQG-CoT~\cite{liang2023prompting} performs chain-of-thought decomposition before question generation. SGSH~\cite{guo2024sgsh} predicts a question skeleton and uses an LLM to realize the final question. RoleAgentQG~\cite{zhao2024zero} formulates KGQG as a multi-role editorial workflow. R2DQG~\cite{ren2025r2dqg} generates questions through relation-aware decomposition. For all adapted static-KGQG methods, we attach event start and end times to the serialized facts and add the temporal-constraint sequence to the input prompt.

\subsection{Main Results}
\label{sec:main-results}
Table~\ref{tab:main-results} reports the main results on the four \sys-generated benchmark splits. We make three observations. 

\begin{figure}[t!]
    \centering
    \subfigure[ChronoQG-Cron]{
    \begin{minipage}{0.45\linewidth}
        \centering
        \includegraphics[scale=0.26]{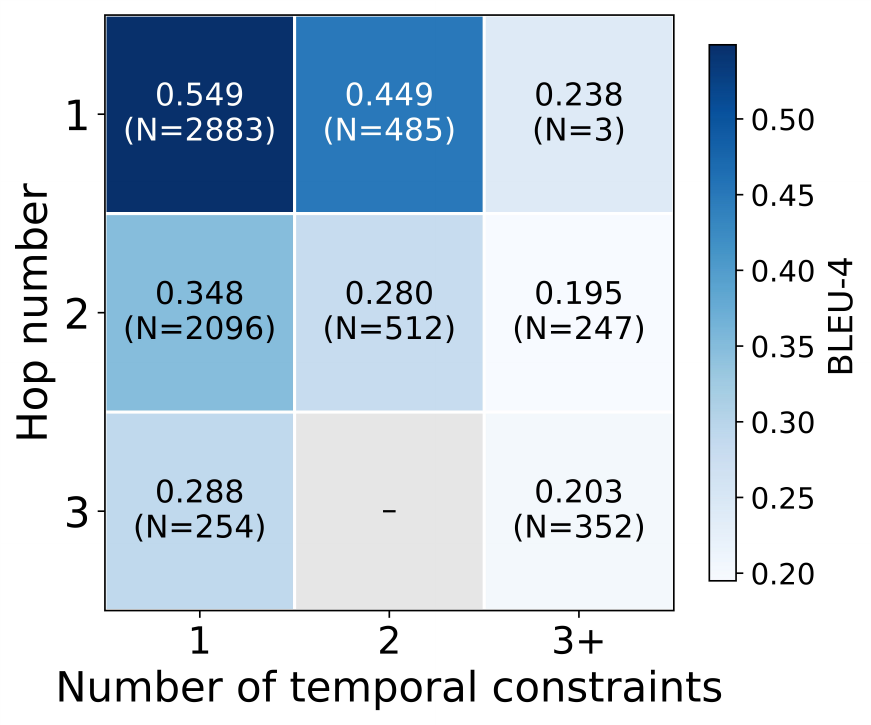}
    \end{minipage}}
    	\subfigure[ChronoQG-Event]{
    \begin{minipage}{0.45\linewidth}
        \centering
        \includegraphics[scale=0.26]{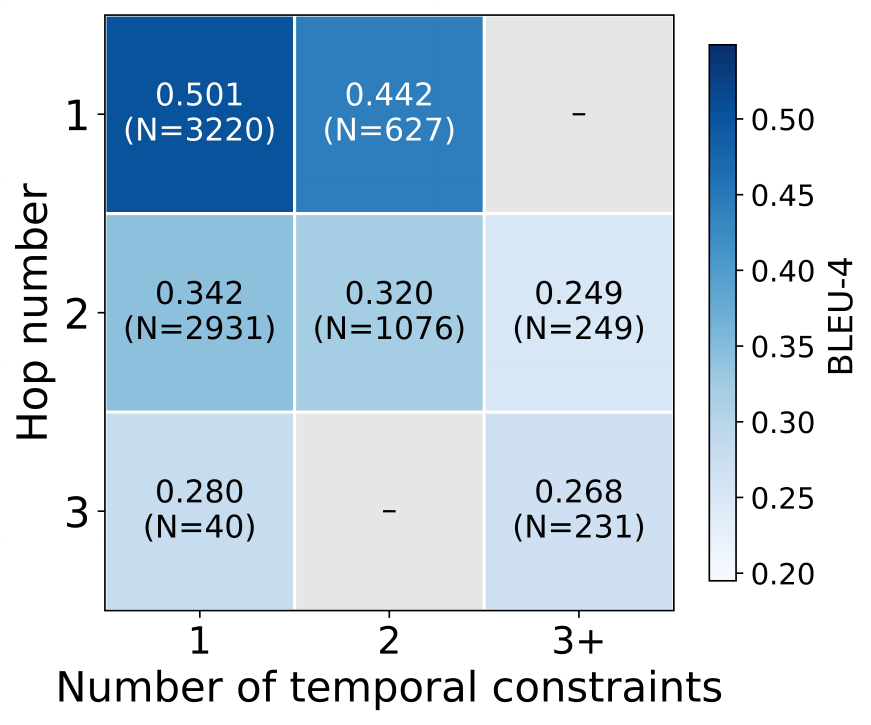}
    \end{minipage}}
    \caption{A heatmap illustrating performance variations with respect to hop count and temporal-constraint count under the zero-shot prompting strategy using GPT-4o-mini. Cases with temporal constraint number larger than 2 are merged.}
    \label{fig:heat-map}
\end{figure}

\begin{table}[t!]
\centering
\footnotesize
\caption{EventKG results of adapted LLM-based KGQG methods with different LLM backbones.}
\label{tab:eventkg-inference-backbone}
\renewcommand{\arraystretch}{1.08}
\setlength{\tabcolsep}{4.0pt}
\resizebox{\columnwidth}{!}{
\begin{tabular}{llccc|ccc}
\toprule
\multirow{2}{*}{\textbf{Method}}
& \multirow{2}{*}{\textbf{Backbone}}
& \multicolumn{3}{c|}{\textbf{Event-S}}
& \multicolumn{3}{c}{\textbf{Event-M}} \\
\cmidrule(lr){3-5} \cmidrule(l){6-8}
& & BLEU-4 & ROUGE-2 & ROUGE-L
& BLEU-4 & ROUGE-2 & ROUGE-L \\
\midrule
\multirow{2}{*}{SGSH}
& GPT-4o-mini & 0.263 & 0.403 & 0.508 & 0.306 & 0.425 & 0.520 \\
& GPT-4o      & 0.320 & 0.469 & 0.579 & 0.308 & 0.434 & 0.536 \\
\midrule
\multirow{2}{*}{KQG-CoT}
& GPT-4o-mini & 0.249 & 0.396 & 0.502 & 0.188 & 0.334 & 0.449 \\
& GPT-4o      & 0.318 & 0.460 & 0.570 & 0.257 & 0.399 & 0.499 \\
\midrule
\multirow{2}{*}{RoleAgentQG}
& GPT-4o-mini & 0.241 & 0.383 & 0.508 & 0.169 & 0.300 & 0.418 \\
& GPT-4o      & 0.226 & 0.377 & 0.504 & 0.193 & 0.334 & 0.450 \\
\bottomrule
\end{tabular}
}
\end{table}

\noindent\textbf{Observation 1.} LLM prompting baselines generally outperform adapted LLM-based static methods. Across the four splits, zero-shot, zero-shot CoT, and few-shot prompting with general-purpose LLMs usually achieve higher BLEU-4, ROUGE-2, and ROUGE-L scores than KQG-CoT, SGSH, RoleAgentQG, and R2DQG. This strongly suggests that explicitly providing the temporal subgraph, target answer, reasoning path, and temporal constraints is a strong formulation for TKGQG. Among prompting baselines, closed-source LLMs tend to perform better than open-source LLMs. For example, GPT-4o-mini, GPT-4o, and Gemini-3.1-pro-preview consistently achieve competitive results across the four splits, while DeepSeek-V3, GLM-4.6, and Qwen2.5-32B show larger performance variation. In contrast, adapted static KGQG methods are generally less effective because their skeletons, decomposition strategies, or editorial workflows are designed for timeless graph triples and do not explicitly model temporal validity, temporal-constraint types, or their attachment to specific graph relations.

\noindent\textbf{Observation 2.} Multi-constraint splits are consistently harder than single-constraint splits. Across most backbones and methods, performance drops from the `-S'' splits to the corresponding `-M'' splits. For instance, GPT-4o-mini with zero-shot prompting drops from 0.458 to 0.288 BLEU-4 on Cron and from 0.410 to 0.328 BLEU-4 on Event. Similar degradation is also observed for ROUGE-2 and ROUGE-L. This confirms that temporal-constraint count is a meaningful difficulty dimension: adding temporal constraints requires the model to preserve more fine-grained temporal semantics and to bind them correctly to the support subgraph.

\noindent\textbf{Observation 3.} Even the strongest methods still leave substantial room for improvement. Although full-training KGQG methods such as G2S-AE-RL achieve the best scores in several settings, their performance remains far from saturated, especially on the multi-constraint splits. Similarly, the best LLM prompting baselines still obtain limited BLEU-4 scores on Cron-M and Event-M, indicating that current LLMs do not reliably preserve answer-determining temporal constraints. Overall, these results demonstrate that \sys exposes a clear gap between static KGQG and TKGQG: existing methods, whether prompt-based, adapted from static KGQG, or fully trained, still struggle to generate questions that are faithful to both topology and temporal constraints.

\subsection{Benchmark Analysis}
\label{sec:analysis}

\subsubsection{Topology-temporal influence}
\label{sec:analysis-structure}

Figure~\ref{fig:heat-map} analyzes how two factors, graph hop count and temporal-constraint count, affect generation difficulty under zero-shot prompting with GPT-4o-mini. We first examine the effect of hop count. When the temporal-constraint count is fixed, increasing the hop count generally leads to lower BLEU-4 scores. On ChronoQG-Cron, under one time constraint, performance drops from 0.547 for one-hop questions to 0.356 for two-hop questions and 0.307 for three-hop questions. A similar pattern appears on ChronoQG-Event, where the score decreases from 0.500 to 0.320 and then to 0.278. This shows that longer reasoning chains make question generation harder, since the model must preserve more relations and bind the target answer to a larger support structure.

We then further examine the effect of temporal-constraint count. When the hop count is fixed, adding more temporal constraints also reduces performance. On ChronoQG-Cron, one-hop questions drop from 0.547 BLEU-4 with one temporal constraint to 0.443 with two temporal constraints, and further to 0.238 with three or more temporal constraints. The same trend holds for two-hop questions, where the score decreases from 0.356 to 0.275 and then to 0.185. ChronoQG-Event shows a consistent but milder trend: for two-hop questions, BLEU-4 decreases from 0.320 with one temporal constraint to 0.310 with two temporal constraints and 0.233 with three or more temporal constraints. These results indicate that temporal constraints introduce an additional source of difficulty beyond graph topology, because the model must not only accurately verbalize the support subgraph but also preserve the temporal conditions attached to specific facts. 

\begin{figure*}[t!]
  \centering
  \includegraphics[scale=0.3]{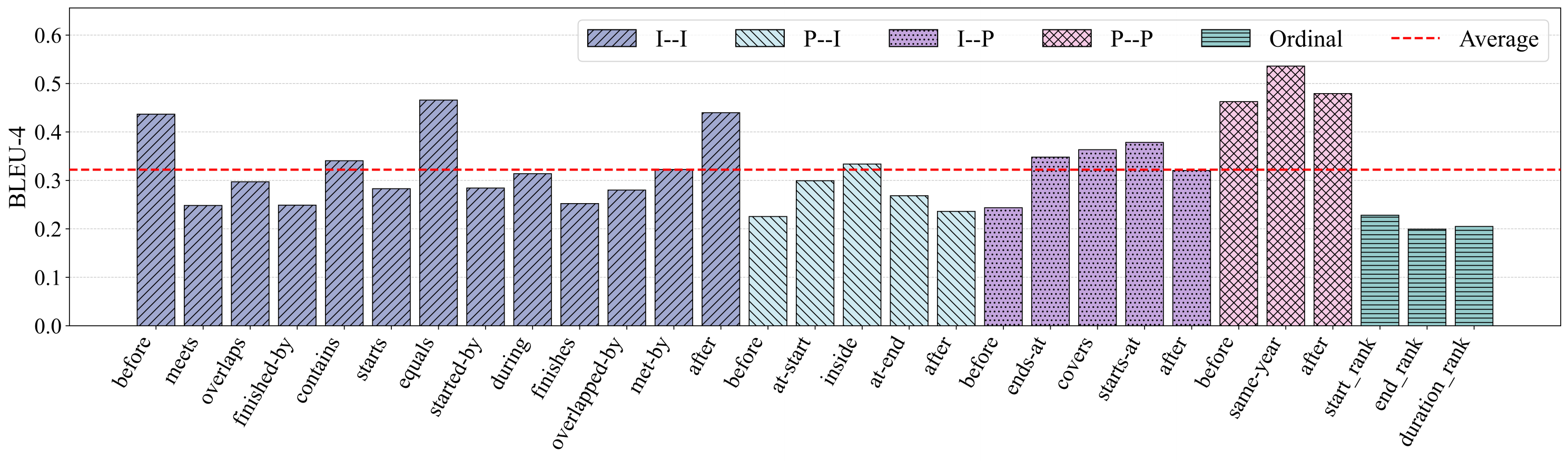}
    \caption{BLEU-4 by temporal-relation type. Scores are aggregated over both source TKGs and the single-/multi-constraint splits. Colors indicate temporal-relation families.}
  \label{fig:category-bars}
\end{figure*}

\subsubsection{Backbone influence}
Table~\ref{tab:eventkg-inference-backbone} examines whether the weakness of adapted static KGQG methods can be mitigated by using a stronger LLM backbone. Replacing GPT-4o-mini with GPT-4o generally improves SGSH and KQG-CoT on both Event-S and Event-M. For example, KQG-CoT increases from 0.249 to 0.318 BLEU-4 on Event-S and from 0.188 to 0.257 on Event-M. SGSH also improves on Event-S and achieves a slight gain on Event-M. However, the overall performance remains unsatisfactory. Even with GPT-4o, these adapted KGQG methods still do not consistently surpass the simple prompting baselines reported in Table~\ref{tab:main-results}, and their scores remain especially limited on the multi-constraint split.

This result suggests that the main bottleneck is not merely the language modeling capability of the backbone, but the mismatch between static KGQG method designs and the requirements of TKGQG. Methods such as SGSH, KQG-CoT, and RoleAgentQG were originally designed for timeless graph triples, where question generation mainly requires preserving graph structure and answer grounding. In TKGQG, however, the model must additionally preserve temporal validity, temporal-constraint types, and the attachment between each temporal constraint and its corresponding graph fact. Simply replacing the backbone with a stronger LLM improves surface realization, but does not fundamentally solve this topology-temporal binding problem. Therefore, the results further confirm that TKGQG requires methods specifically designed for temporally constrained graph evidence, rather than direct adaptations of static KGQG pipelines.

\begin{table}[t!]
\centering
\small
\caption{API cost in USD for running all four datasets.}
\label{tab:cost}
\renewcommand{\arraystretch}{1.08}
\setlength{\tabcolsep}{6.0pt}
\resizebox{0.72\columnwidth}{!}{%
\begin{tabular}{l|cc}
\toprule
\textbf{Method} & \textbf{GPT-4o-mini} & \textbf{GPT-4o} \\
\midrule
Zero-shot      & 2.30  & 38.34  \\
Zero-shot CoT         & 2.45  & 40.79  \\
Few-shot    & 6.73  & 112.17 \\
\midrule
SGSH        & 17.16 & 286.03 \\
KQG-CoT     & 23.15 & 385.84 \\
RoleAgentQG & 20.94 & 349.10 \\
\bottomrule
\end{tabular}%
}
\end{table}

\subsubsection{Model cost}
\label{sec:analysis-backbone}
Table~\ref{tab:cost} compares the estimated API cost of different generation strategies. Existing LLM-based KGQG methods incur substantially higher costs than simple prompting baselines. On GPT-4o-mini, zero-shot prompting costs only 2.30 for all four released datasets, while SGSH, KQG-CoT, and RoleAgentQG require 17.16, 23.15, and 20.94, respectively. The gap becomes even larger with GPT-4o, where zero-shot prompting costs 38.34, but the adapted KGQG methods cost between 286.03 and 385.84. This overhead mainly comes from their multi-stage designs, such as skeleton prediction, decomposition, and multi-role editing, which require multiple LLM calls for each instance. However, as shown in Table~\ref{tab:main-results}, these additional calls do not translate into better TKGQG performance. In contrast, direct LLM prompting achieves stronger results with much lower cost by explicitly conditioning on the temporal subgraph, target answer, reasoning path, and temporal constraints in a single generation step. These results suggest that current KGQG pipelines are not only less effective for TKGQG, but also less cost-efficient than simpler LLM prompting formulations.

\begin{table}[t!]
\centering
\small
\caption{Human evaluation of generated questions. Fluency measures linguistic naturalness, faithfulness measures whether the question preserves the given graph and temporal constraints, and answerability measures whether the question can be answered from the provided evidence.}
\label{tab:human-evaluation}
\renewcommand{\arraystretch}{1.08}
\setlength{\tabcolsep}{5.5pt}
\begin{tabular}{l|ccc}
\toprule
\textbf{Dataset} & \textbf{Fluency} & \textbf{Faithfulness} & \textbf{Answerability} \\
\midrule
Cron-S      & 4.85 & 4.83 &  4.86\\
Cron-M       & 4.71 & 4.81 &  4.65\\
\midrule
Event-S      & 4.84 & 4.71 &  4.85\\
Event-M     & 4.59 & 4.69 &  4.72\\
\bottomrule
\end{tabular}
\end{table}

\begin{figure*}[t!]
\centering
\includegraphics[width=\textwidth]{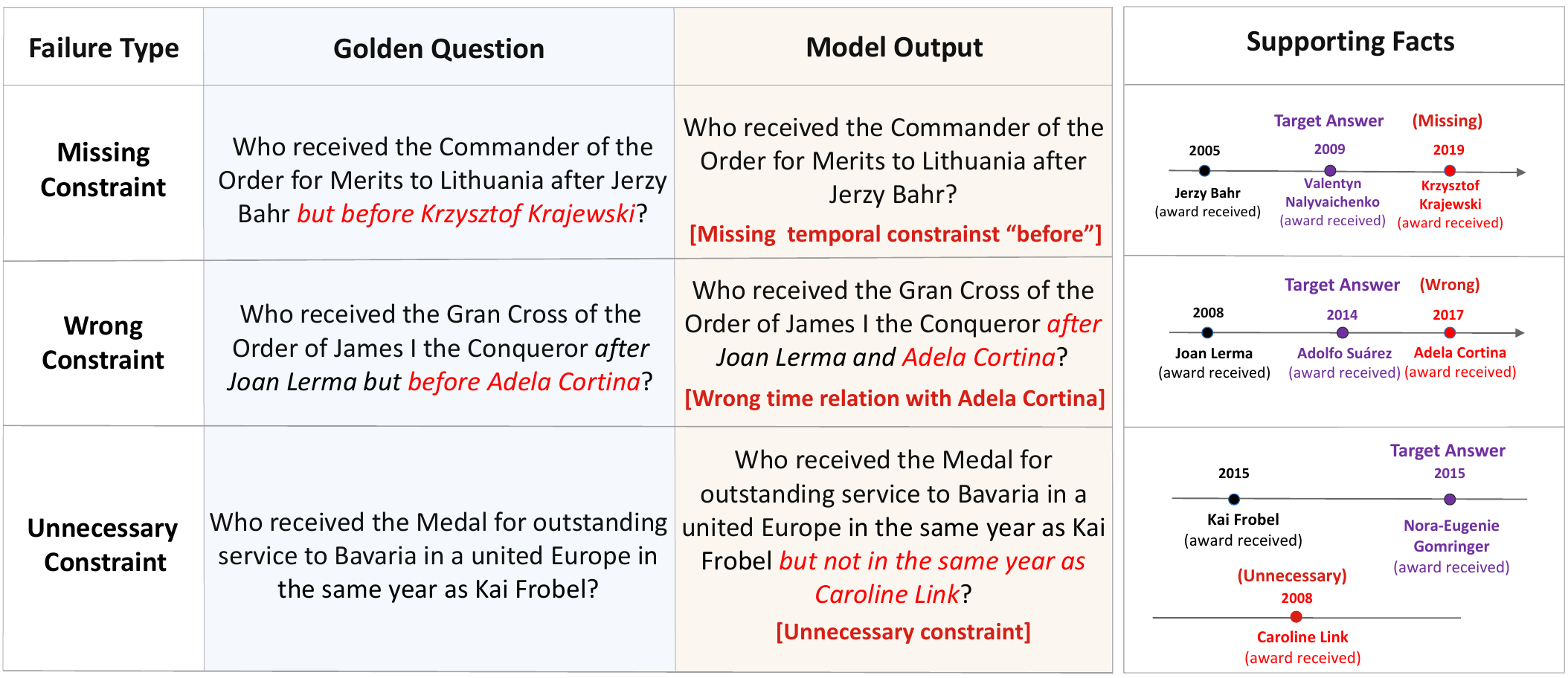}
\caption{Failure cases of TKGQG.}
\label{fig:failure-case}
\end{figure*}

\subsubsection{Temporal taxonomy.}
Figure~\ref{fig:category-bars} analyzes performance across different temporal-constraint types. The results show that generation difficulty varies substantially across constraint families, indicating that TKGQG cannot be fully characterized by hop count or temporal-constraint count alone. Interval--interval constraints exhibit large internal variation: boundary-sensitive relations such as \emph{meets}, \emph{starts}, and \emph{finishes} are generally harder than more intuitive ordering or containment relations, because the model must verbalize precise interval boundaries rather than coarse temporal order. Point--interval and interval--point constraints are also challenging, especially when the generated question needs to attach a point event to the start, end, or interior of an interval-valued fact. In contrast, point--point constraints are relatively easier, since they often correspond to familiar temporal expressions such as before, after, or same year. Ordinal constraints show consistently lower performance, suggesting that ranking-based temporal reasoning is difficult for current models: the model must not only express temporal order, but also preserve the selection criterion over a candidate set. Overall, the results demonstrate that temporal-constraint type is an important diagnostic dimension of \sys. Different temporal operators impose different realization and grounding requirements, and existing models remain sensitive to these fine-grained temporal semantics.

\subsubsection{Human evaluation.}
To assess the quality of the generated questions, we conduct a human evaluation on four dataset splits. For each dataset, we randomly sample 500 generated questions from the final verified set. The sampled questions are evaluated by human annotators with background knowledge in knowledge graphs and question answering, so that they can judge not only surface fluency but also whether the generated question is consistent with the underlying graph evidence and temporal constraints. Each annotator is provided with the generated question, the target answer, the supporting facts, and the corresponding temporal constraints. We ask the annotators to rate each question along three dimensions using a 1--5 Likert scale, where 5 indicates the best quality. Fluency measures whether the question is grammatically correct and linguistically natural. Faithfulness measures whether the question preserves the given graph structure, supporting facts, target answer, and temporal constraints. Answerability measures whether the question can be answered from the provided evidence without requiring external knowledge. The final score for each dimension is computed as the average over the 500 sampled questions.

As shown in Table~\ref{tab:human-evaluation}, the generated questions achieve consistently high scores across all four datasets. Fluency scores are above 4.69 for all splits, indicating that the rewriting process produces natural and readable questions. Faithfulness scores are also high, showing that most generated questions preserve the intended graph and temporal semantics. Answerability remains high across all datasets, demonstrating that the generated questions can generally be answered using the provided supporting facts. Overall, the human evaluation confirms that our framework produces questions that are not only fluent, but also faithful to temporal KG evidence and answerable from the given context.

\begin{table}[t!]
\centering
\small
\caption{Comparison between raw and final questions across four datasets. $\Delta_{\mathrm{len}}$ denotes the relative reduction in question length after rewriting. BLEU-4 is computed between raw and final questions to quantify surface-form divergence. Preservation measures the percentage of entities in the raw question that are retained in the final question.}
\label{tab:rewrite-comparison}
\renewcommand{\arraystretch}{1.08}
\setlength{\tabcolsep}{6pt}
\begin{tabular}{l|ccc}
\toprule
\textbf{Dataset} 
& \textbf{$\Delta_{\mathrm{len}}$ (\%)} 
& \textbf{BLEU-4} 
& \textbf{Preservation (\%)} \\
\midrule
Cron-S  & 49.8 & 0.10 & 98.0 \\
Cron-M  & 62.1 & 0.08 & 94.4  \\
Event-S & 52.6 & 0.09 & 94.1  \\
Event-M & 63.1 & 0.06 & 92.1  \\
\midrule
\textbf{Avg.}    & 59.6 & 0.08 & 94.7 \\
\bottomrule
\end{tabular}
\end{table}

\subsubsection{Rewriting performance.}
Table~\ref{tab:rewrite-comparison} compares the raw template questions with the final rewritten questions across four datasets. Here, $\Delta_{\mathrm{len}}$ measures the relative reduction in question length after rewriting, and Preservation measures the percentage of entities appearing in the raw question that are retained in the rewritten question. Overall, the rewriting process simplifies the original template-style questions, with an average length reduction of 56.9\%. The reduction is consistent across all datasets, ranging from 49.8\% on Cron-S to 63.1\% on Event-M, showing that the rewriting step effectively removes rigid and verbose template expressions. Despite this large compression, the rewritten questions preserve most of the semantic content. Entity preservation remains above 92\% for all datasets, with an average of 94.7\%, indicating that the rewriting process largely retains the entities required to express the original graph and temporal constraints. Meanwhile, the BLEU-4 scores between raw and rewritten questions are low, with an average of 0.083. This does not indicate poor quality; rather, it shows that the final questions are not lexical copies or minor edits of the templates. Instead, the model substantially changes the surface form, producing more natural questions while preserving the key entities and intended semantics. These results suggest that our rewriting process achieves high-quality naturalization: it greatly simplifies the expression, departs from template-like wording, and still maintains the essential semantic content of the original question.

\subsubsection{Constraint contribution}
Table~\ref{tab:root-seed} reports the average number of root-seed candidates before progressive constraint application. Across the four datasets, the initial candidate sets are non-trivial, ranging from 5.34 to 20.30 on average, indicating that the root relation alone usually cannot determine a unique answer. This confirms the necessity of progressive sampling: starting from a relatively broad candidate set, our method incrementally applies structural hops and temporal constraints, and each accepted constraint must reduce the remaining candidates. As a result, the sampling process effectively turns ambiguous root seeds into well-constrained question instances. The larger candidate sets in multi-hop settings, especially Event-S at 3-hop, further show that graph traversal can expand the search space, making progressive temporal filtering essential for controlling answer ambiguity and producing questions with unique, verifiable answers.

\subsection{Case Study}
Figure~\ref{fig:failure-case} presents three representative failure cases in TKGQG. Although the generated questions are generally fluent, they may fail to preserve the exact temporal semantics encoded in the template. The first case shows a missing-constraint error: the model removes the ``before'' condition and produces a question whose answer set is no longer restricted by the required temporal order. The second case illustrates a wrong-constraint error, where the model retains a temporal expression but changes its relation from the intended ``after Joan Lerma but before Adela Cortina'' to a different condition involving both entities, leading to an incorrect target answer. The third case shows an unnecessary-constraint error: the model introduces an additional temporal comparison that is not present in the gold question, thereby introducing a redundant constraint.

These cases suggest that the main challenge is not surface-level fluency, but faithful preservation of temporal constraints during natural language realization. Even when the generated question appears grammatical and semantically plausible, small changes to temporal operators can alter the denotation of the question. This observation motivates the use of explicit verification in our framework: by checking whether the rewritten question preserves the intended supporting facts and target answer, the verifier can identify questions that are fluent but temporally inconsistent. The case study therefore complements the quantitative results by showing that temporal-relation fidelity is a critical factor for reliable TKGQG.

\begin{table}[t!]\small
\centering
\caption{Average number of root-seed candidates before progressive constraint application.}
\label{tab:root-seed}
\renewcommand{\arraystretch}{1.08}
\setlength{\tabcolsep}{7.0pt}
{
\begin{tabular}{l|ccc}
\toprule
\multirow{2}{*}{\textbf{Dataset}}
& \multicolumn{3}{c}{\textbf{\# Root Seed}}\\
\cmidrule(lr){2-4} 
& 1-Hop & 2-Hop & 3-Hop\\
\midrule
Cron-S & 12.93 & 11.79 & 14.01\\
Cron-M & 13.18 & 11.36 & 9.04\\
\midrule
Event-S & 5.56 & 13.11 & 20.30\\
Event-M & 5.34 & 12.17 & 9.52\\
\bottomrule
\end{tabular}
}
\end{table}

\subsection{LLM Prompt}
\begin{figure}[t!]
    \centering
    \includegraphics[width=0.8\linewidth]{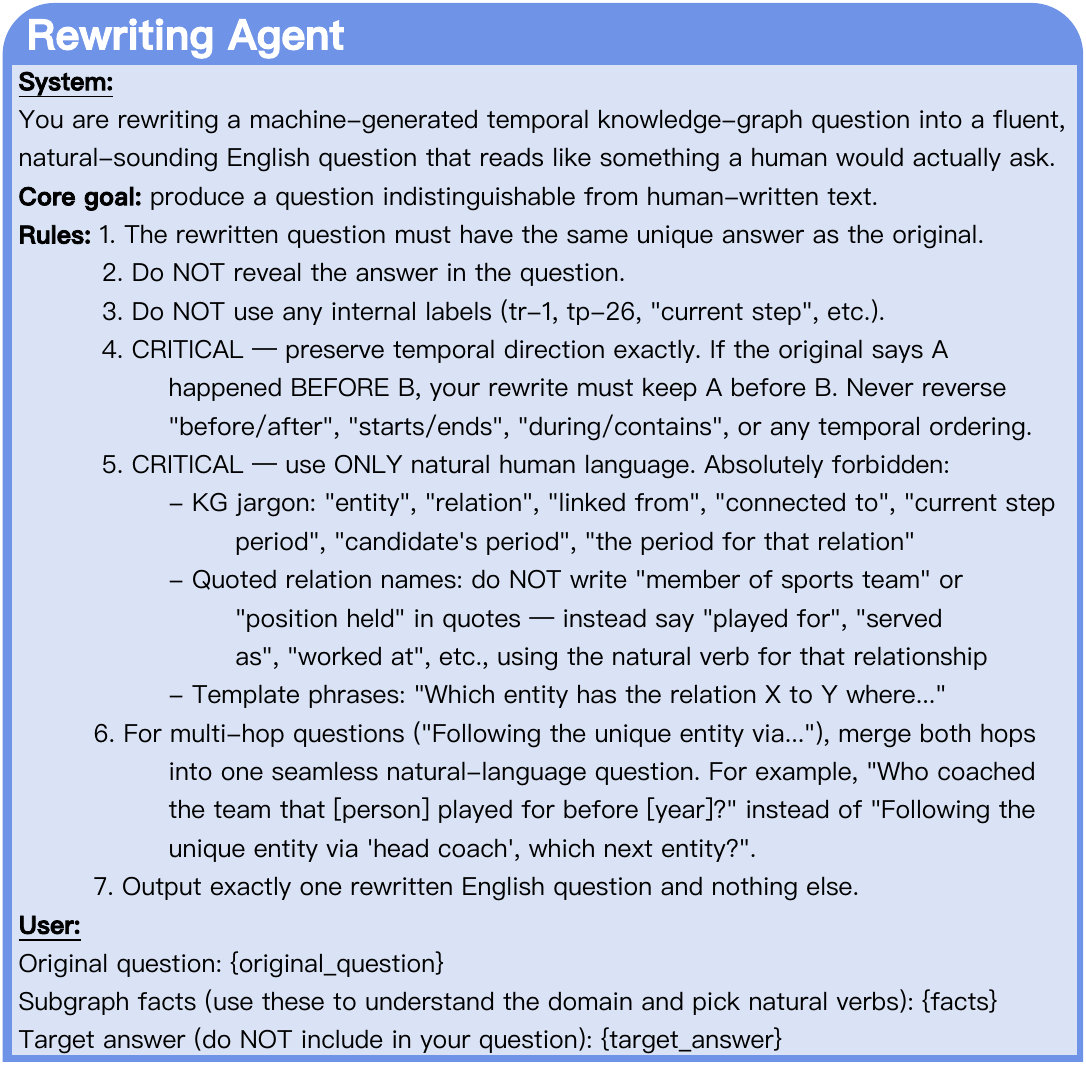}
    \caption{Prompt of rewriting agent.}
    \label{fig:fig8}
\end{figure}

\begin{figure}[t!]
    \centering
    \includegraphics[width=0.8\linewidth]{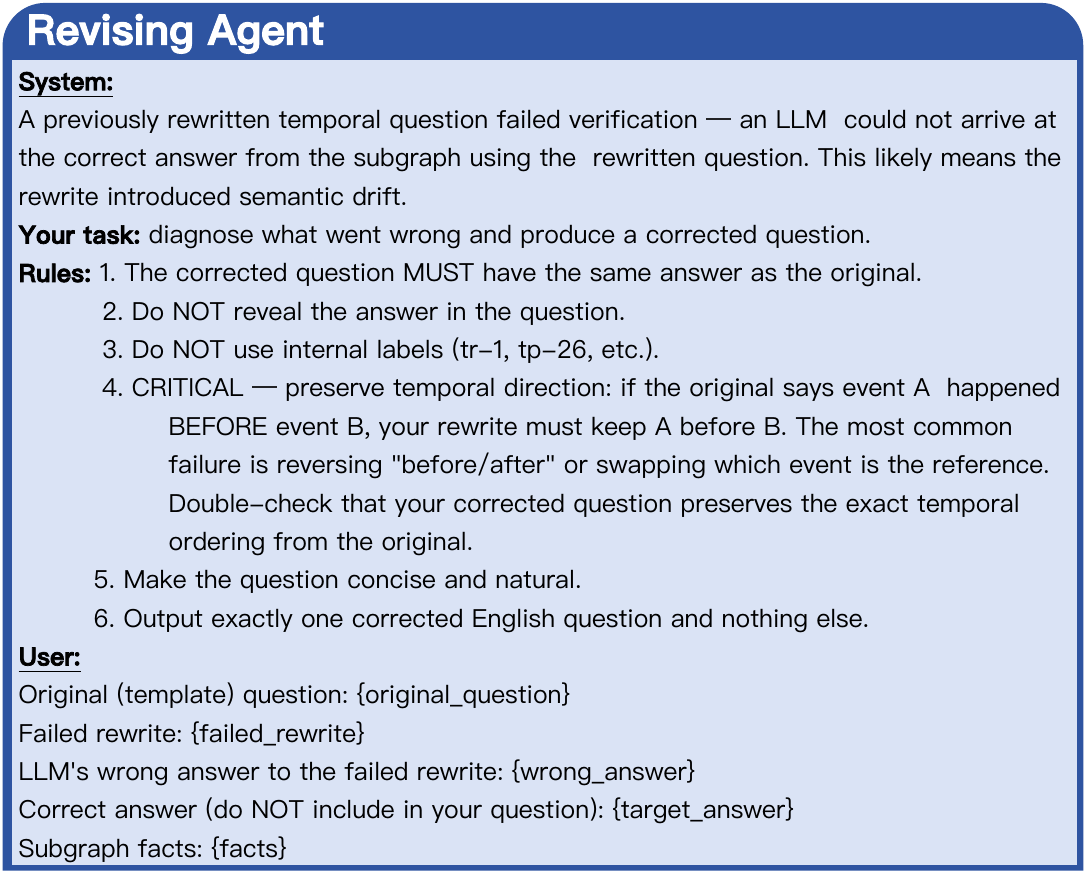}
    \caption{Prompt of revising agent.}
    \label{fig:fig9}
    \vspace{-1em}
\end{figure}

Figures~\ref{fig:fig8} and ~\ref{fig:fig9} show the prompts of the rewriting and revising agents, respectively. The rewriting prompt converts template questions into fluent natural-language questions while preserving the original answer and temporal meaning. The revising prompt is used to correct failed rewrites by eliminating semantic drift, especially temporal-order errors, and producing a verified natural question.

\section{Conclusion}
\label{sec:conclusion}
In this paper, we present \textsc{ChronoQG}, a framework for constructing temporally expressive and hop-bounded TKGQG benchmarks. To address the limited temporal coverage and weak constraint control of existing KGQG datasets, \textsc{ChronoQG} introduces a comprehensive temporal-relation taxonomy, a topology--temporal subgraph sampling strategy, and a verifier-guided rewriting pipeline to generate fluent questions that preserve the intended graph and temporal semantics.
Using this framework, we construct four benchmark splits from two source TKGs, resulting in 16,011 verified questions with diverse hop structures and temporal constraint patterns. Extensive analyses show that current methods, including both LLM-based approaches and adapted static KGQG baselines, still struggle with temporally constrained question generation, especially as topological complexity and temporal-constraint complexity increase. This reveals a clear gap between existing KGQG techniques and the requirements of temporally faithful question generation. We hope \textsc{ChronoQG} serves as a useful benchmark for exposing this gap and advancing models that reason over complex temporal knowledge.

\bibliographystyle{ACM-Reference-Format}
\bibliography{sample}

\end{document}